\def\year{2020}\relax
\documentclass[letterpaper]{article} 
\usepackage{aaai20}  
\usepackage{times}  
\usepackage{helvet} 
\usepackage{courier}  
\usepackage[hyphens]{url}  
\usepackage{graphicx} 
\usepackage{inputenc} 
\usepackage{booktabs}       
\usepackage{amsfonts}       
\usepackage{nicefrac}       
\usepackage{microtype}      
\usepackage{subfigure}
\usepackage{amssymb}
\usepackage{amsmath}
\newcommand{\Norm}[1]{\left\lVert#1\right\rVert}
\newcommand{\norm}[1]{\left\lvert#1\right\rvert}
\DeclareMathOperator*{\argmax}{arg\,max}
\usepackage{stmaryrd}

\usepackage{pifont}
\usepackage{multirow}
\usepackage{bm}
\usepackage[english]{babel}
\newtheorem{prop}{Proposition}

\usepackage{cuted}
\usepackage{adjustbox}
\newcommand{\ie}{\mbox{\it{i.e.,\ }}}
\newcommand{\eg}{\mbox{\it{e.g.,\ }}}
\newcommand{\para}[1]{\vspace*{1ex}\noindent\textbf{#1} }

\title{LMLFM: Longitudinal Multi-Level Factorization Machine}
\author{
Junjie Liang,\textsuperscript{\rm1,\rm2}
Dongkuan Xu,\textsuperscript{\rm2}
Yiwei Sun,\textsuperscript{\rm1,\rm3}
Vasant Honavar\textsuperscript{\rm1,\rm2,\rm3,\rm4}\\
\textsuperscript{\rm1} Artificial Intelligence Research Laboratory, Pennsylvania State University\\
\textsuperscript{\rm2}College of Information Sciences and Technology, Pennsylvania State University\\
\textsuperscript{\rm3} Department of Computer Science and Engineering, Pennsylvania State University\\
\textsuperscript{\rm4} Institute of Computational and Data Sciences, Pennsylvania State University\\
\{jul672,dux19,vhonavar\}@ist.psu.edu, yus162@psu.edu
}

\begin{document}

\maketitle

\begin{abstract}
We consider the problem of learning predictive models from longitudinal data, consisting of irregularly repeated, sparse observations from a set of individuals over time. Such data often exhibit {\em longitudinal correlation} (LC) (correlations among observations for each individual over time), {\em cluster correlation} (CC) (correlations among individuals that have similar characteristics), or both. These correlations are often accounted for using  {\em mixed effects models} that include {\em fixed effects} and {\em random effects}, where the fixed effects capture the regression parameters that are shared by all individuals, whereas random effects capture those parameters that vary across individuals. However, the current state-of-the-art methods are unable to select the most predictive fixed effects and random effects from a large number of variables, while accounting for complex correlation structure in the data and non-linear interactions among the variables. We propose Longitudinal Multi-Level Factorization Machine (LMLFM), to the best of our knowledge, the first model to address these challenges in learning predictive models from longitudinal data. We establish the convergence properties, and analyze the computational complexity, of LMLFM. We present  
results of experiments with both simulated and real-world longitudinal data which show that LMLFM outperforms the state-of-the-art methods in terms of  predictive accuracy, variable selection ability, and scalability to data with large number of variables. The code and supplemental material is available at \url{https://github.com/junjieliang672/LMLFM}.
\end{abstract}

\section{Introduction}
\label{sec:intro}
\begin{figure*}[tb]
\centering
\includegraphics[width=.85\textwidth]{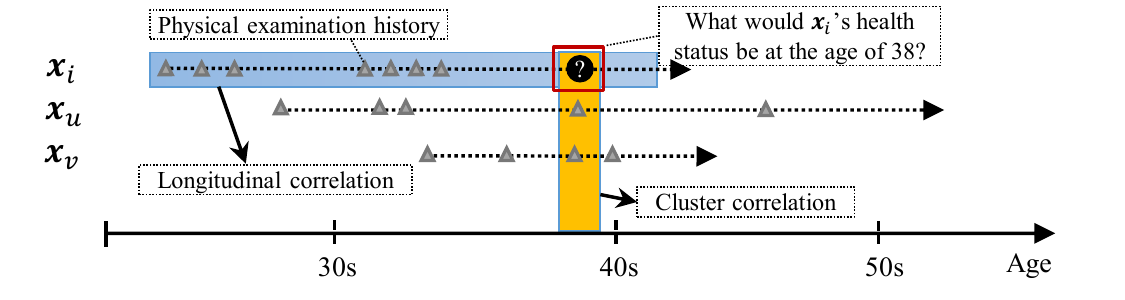}
\caption{An example of longitudinal data. Note that the type of correlation behind the data is unknown. Hence, a predictive model should not only learn to accurately predict the outcome, but also exploit and identify the type of data correlation.}
\label{fig:fig1}
\end{figure*}

{\em Longitudinal data} consist of repeated observations from a set of individuals over time. Such data are common in many areas, including health sciences, social sciences and economics. Consider for example, the scenario shown in Fig.~\ref{fig:fig1}. To predict an individual $\bm{x}_i$'s health status at the age of 38, we should take into account both $\bm{x}_i$'s history of physical examinations, as well as those of other individuals who are similar to $\bm{x}_i$ in age and other characteristics. Clearly, such longitudinal data often exhibit {\em longitudinal correlation} (LC) (correlations among observations for each individual over time), {\em cluster correlation} (CC) (correlations among individuals that have similar characteristics), or both (multi-level correlation) \cite{finch2016multilevel}, and hence are not independent and identically distributed. Analysis that does not account for such correlations can lead to misleading statistical inferences \cite{gibbons1997random}. 
To account for data correlations, state-of-the-art longitudinal data analysis  \cite{gibbons1997random,lozano2012multi,groll2014variable} often relies on {\em mixed effects models} that include {\em fixed effects} and {\em random effects}, where the fixed effects capture the regression parameters that are shared by all individuals, whereas random effects capture those parameters that vary across individuals. In practice, the design of mixed effects models relies on expert input to decide which variables are subject to random effects as opposed to fixed effects, or a process of  trial and error. However, existing mixed effects models are very computationally intensive, with the computational cost scaling with $O(q^3)$, where $q$ is the number of variables that are subject to random effects which limits their applicability to relatively low-dimensional data. While with the advent of big data, variants of dimensionality reduction methods such as LASSO have been explored in the longitudinal setting  \cite{schelldorfer2011estimation,lozano2012multi,groll2014variable,xu2015longitudinal,ratkovic2017sparse,marino2017covariate,lu2017multilevel,finch2018modeling}, most such methods are limited to selecting only fixed effects. There is limited work on jointly selecting fixed and random effects, e.g., penalized likelihood methods \cite{ibrahim2011fixed,bondell2010joint,hui2017joint} and Bayesian models \cite{chen2003random,yang2019bayesian}. However, their applicability is limited by their high computational cost and reliance on linear models.

\para{Contributions.}
This paper aims to address the urgent need for effective models that can handle high-dimensional longitudinal data where the number of variables is very large compared to the size of the population, the interactions among variables can be nonlinear, the fixed and random effects are a priori unspecified, and the data exhibit correlation structure (LC, CC, or both).  Specifically, we introduce Longitudinal Multi-Level Factorization Machine (LMLFM), a novel, efficient, provably convergent  extension of {\em Factorization Machines} (FM) \cite{rendle2012factorization} for predictive modeling of high-dimensional longitudinal data. LMLFM inherit the advantages of FM, e.g., the  ability to reliably estimate the model parameters from high-dimensional data and model non-linear interactions. Further, LMLFM can automatically select fixed and random effects even in the presence of multi-level correlation, and greatly reduce the need for hyper-parameter tuning using a novel hierarchical Bayesian formulation. Specifically, LMLFM adopts two layers of Laplace prior, one for sparsifying the latent representation and one for identifying fixed effects and random effects. We solve the LMLFM using the iterated conditional modes (ICM) algorithm \cite{besag1986statistical} which offers efficient optimization with strong convergence guarantees. Experimental results with simulated data show that LMLFM can readily handle longitudinal data with over 5000 variables whereas the existing mixed effects models fail when the number of variables exceeds 100. Experiments with two real-world data sets show that LMLFM (i) compares favorably with the state-of-the-art baselines in terms of predictive accuracy; (ii) yields sparse and easy-to-interpret predictive models; and (iii) effectively selects the relevant variables, which are consistent with the published findings \cite{bromberger2011mood,dolan2008we}. 

\section{Related Work}
\label{sec:related}
Popular longitudinal data analysis methods include Generalized Estimating Equations (GEE) \cite{liang1986longitudinal} and Generalized Mixed Effects Models (GMEM) \cite{fitzmaurice2012applied}. GEE are marginal models which only estimate the average outcome (or fixed effects) over the population \cite{liang1986longitudinal}. In contrast, GMEM are conditional models that provide the expectation of the conditional distribution of the outcome given the random effects. 

There is much interest in the problem of variable selection in longitudinal data  \cite{schelldorfer2011estimation,groll2014variable}. Existing techniques focus on selecting only the fixed effects, under the assumption that the type of correlation is correctly specified and the random and fixed effects are correctly identified. Their high computational cost limits their applicability to data with small numbers of variables \cite{chen2018latent}. There is limited work on the more challenging problem of selecting both fixed effects and random effects. Existing methods typically rely on adding a sparsity inducing penalty, e.g., LASSO or its variants, to the GMEM objective function \cite{bondell2010joint,ibrahim2011fixed,muller2013model,hui2017hierarchical,hui2017joint}. While Bayesian methods, \eg \cite{chen2003random,yang2019bayesian}, offer a conceptually attractive alternative  to  penalized likelihood methods for variable selection, they are currently applicable only to 2-level data which exhibit only LC or only CC but not both. Furthermore, most assume a linear mixed model, and hence cannot accommodate non-linear interactions among variables. Because they rely on matrix decomposition and matrix inversion for parameter estimation, their computational complexity is $O(q^3)$, making them unsuitable for high-dimensional longitudinal data.

While there have been a few attempts at applying factorization techniques \cite{zhou2014micro,stamile2017multiparametric,Kidzinski2018LongitudinalDA}, and deep representation learning techniques \cite{xu2019adaptive,xu2019spatio}, their primary focus is to improve the predictive accuracy. These techniques do not explicitly account for the complex correlation structure in the data or distinguish between random effects and fixed effects. In contrast, LMLFM efficiently accounts for complex correlation structure in the data and selects the most predictive fixed and random effects.

\section{Preliminaries}

\para{Notation.}
Scalars are denoted by lowercase letters and vectors by bold lowercase letters. All vectors are column vectors. $\norm{\bm{\theta}}$ refers to the length of $\bm{\theta}$ and $\Norm{\bm{\theta}}_p$ is the $\ell_p$ norm of $\bm{\theta}$.
Matrices are denoted by uppercase letters, \eg $\Theta$ and a set of objects by a bold uppercase letter, \eg $\bm{\Theta}=\{a,\bm{\theta}\}$. 
The calligraphic letters $\mathcal{I}$ and $\mathcal{O}$  denote information related to the individuals and the observations respectively. For example, $\Theta^\mathcal{I}$ refers to the sub-matrix of $\Theta$ associated to the individuals. We use the letters $i,o$ to denote an arbitrary individual and observation respectively. We use $\mathrm{diag}(A)$ to denote the vector of diagonal components of a square matrix $A$.
Because observations occur at  discrete time points, we use observation and time point interchangeably.

\begin{figure}[tb]
\centering
\includegraphics[width=.45\columnwidth] {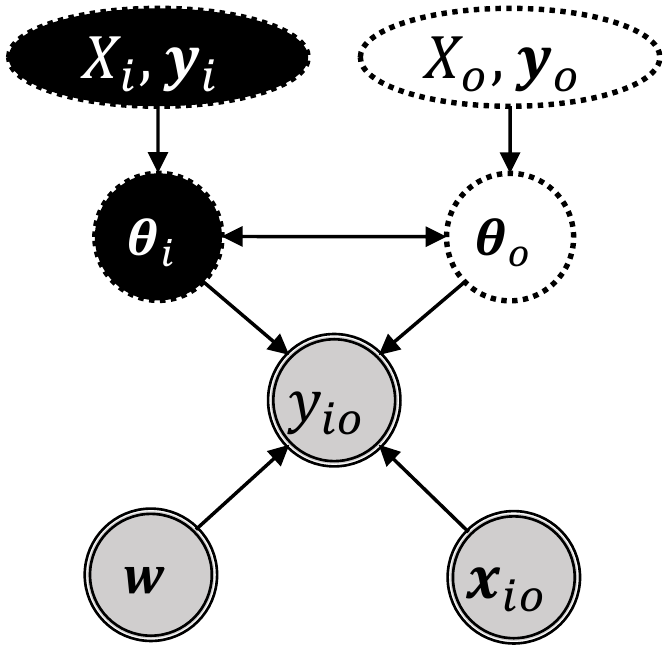}
\caption{Model structure. Sub-graph with grey nodes encodes the structure of standard 1-level regression models;  Sub-graph with grey and black nodes encodes the structure of 2-level LC models; Sub-graph with grey and white nodes denotes the structure of 2-level CC models; and the entire graph encodes the structure of multi-level models. }
\label{fig:assumption}
\end{figure}

\para{Factorization Machines (FM).}
Given a real-valued design feature vector $\bm{x}_{io}\in \mathbb{R}^p$ corresponding to an individual $i$ and observation $o$, factorization machines (FM) \cite{rendle2012factorization} model all nested interactions up to order $d$. For example, the prediction of FM of order $d=2$ is given by:
\begin{equation}
\label{eq1}
\hat{y}_{io}=\bm{x}^\intercal_{io}\bm{w}+\frac{1}{2}\bm{x}^\intercal_{io}M\bm{x}_{io}
\end{equation}
where $M$ is a squared matrix with zeros on the diagonal. The off-diagonal component 
$m_{qt}\in M$ is parameterized as a dot product of two low dimensional embeddings $\bm{\theta}_q,\bm{\theta}_t$. FM can be readily solved by coordinate descent \cite{rendle2012factorization}. The time and space complexity are $O(k\norm{S})$ and $O(\norm{\bm{y}})$ respectively where $\norm{\bm{y}}$ and $\norm{S}$ denote the total number of observations across all individuals, and the data size (\ie $p\norm{\bm{y}}$), respectively.

\para{Linear Mixed Model (LMM).}
We introduce LMM to motivate the design of LMLFM. Let $y_{io}\in\mathbb{R}$ denote the scalar outcome of individual $i=1,\cdots,n$ measured at observation $o=1,\cdots,m$. Let $\bm{x}_{io}\in\mathbb{R}^{p}$, $\bm{z}_{io}\in\mathbb{R}^q\subseteq \bm{x}_{io}$ be variables associated with fixed effects (denoted by $\bm{\beta}$) and random effects (denoted as $\bm{\gamma}_i$) respectively. LMM assumes that the outcome is predicted by\footnote{We omit the error term since it can be readily incorporated into the random effects.}:
\begin{equation}
\label{eq0}
\hat{y}_{io}=\bm{x}^\intercal_{io}\bm{\beta} + \bm{z}^\intercal_{io}\bm{\gamma}_i
\end{equation}
The random effects matrix $\left(\bm{\gamma}_1,...,\bm{\gamma}_n\right)^\intercal$ captures the time-invariant patterns for each individual. For all $i\in \mathcal{I}$, $\bm{\gamma}_i{\sim}N(\bm{0},Q)$, where $Q\in\mathbb{R}^{q\times q}$ is the covariance matrix. The random effects $\bm{\gamma}_i$ serve two purposes: (i) regularizing the effects (similar to the $\ell_2$ norm); and (ii) inducing correlation between the longitudinal observations, \ie $cov(y_{io},y_{ij})=\bm{z}^\intercal_{io}Q\bm{z}_{ij}$.

\section{Longitudinal Multi-Level Factorization Machine (LMLFM)}
\label{sec:LMLFM}
The structures of multi-level models are shown in Fig.~\ref{fig:assumption}. Standard regression models assume that given the variables and regression parameters, the outcomes are i.i.d. and hence yield biased  estimates of parameters in the presence of LC or CC. 2-level models account for the LC or CC by either directly or indirectly specifying a correlation matrix that models the corresponding correlations. Mixed effects models introduce individual (or observation) specific random effects as proxies for the relevant information (see Fig.~\ref{fig:assumption}). A natural approach to extend this design is to incorporate both individual factors $\bm{\theta}_i$ and observation factors $\bm{\theta}_o$, as proxies for the individual and observation specific information.\footnote{It is worth noting that individual/observation factors are distinct from individual/observation random effects, in this work, we use the former to estimate the latter.} The pairwise interactions between such factors in the latent space (see Eq.~\eqref{eq1}) are shown by arrows in Fig.~\ref{fig:assumption}. However, in its current form, the model does not provide a way to relate latent factors to the random effects or to explicitly accommodate complex correlation structures.  \textbf{Given the observed design matrix $X$ and outcomes $\bm{y}$, our goals are to: (i) predict the unknown outcomes $\hat{\bm{y}}$, (ii) jointly select both fixed and random effects and (iii) recover the correlation structure from the data.}
\subsection{Prediction}
\label{sec:prediction}
\begin{figure}[tb]
\centering
\includegraphics[width=.95\columnwidth] {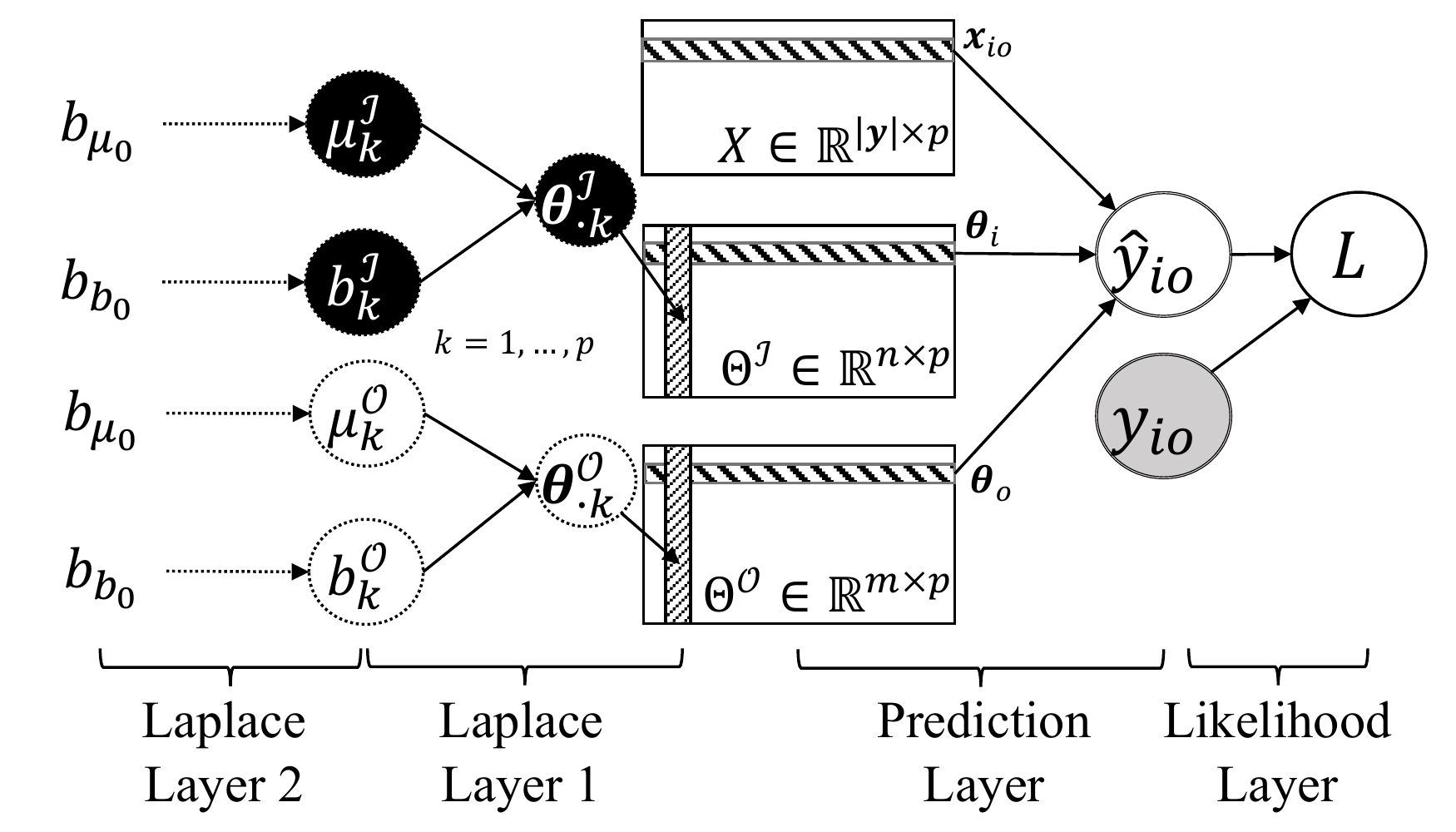}
\caption{The hierarchical Bayesian structure of LMLFM. $L$ stands for the objective function as presented in Eq.~\eqref{eq:map}.}
\label{fig:LMLFM}
\end{figure}

The prediction layer of LMLFM (see Fig.~\ref{fig:LMLFM}) is inspired by both LMM and FM. With a Bayesian framework, all variables are first assumed random. Hence, the LMM prediction in Eq.~\eqref{eq0} reduces to $\hat{y}_{io}=\bm{x}^\intercal_{io}\bm{\gamma}_{io}$ with $\bm{\gamma}_{io}{\sim}N(\bm{\beta},Q)$. Further, to accommodate multi-level correlation, we decompose the random effects as the summation of two subsets of latent factors $\bm{\gamma}_{io}=\bm{\theta}_i+\bm{\theta}_o$, where $\bm{\theta}_i,\bm{\theta}_o$ denote the individual factors and observation factors respectively. Considering the interaction between individual and observation factors, we introduce the following prediction function:
\begin{equation}
\label{eq2}
\hat{y}_{io}=\bm{x}^\intercal_{io}\left(\bm{\theta}_i+\bm{\theta}_o\right)+\bm{\theta}^\intercal_i\bm{\theta}_o
\end{equation}
Recall that in FM, the design feature vector $\bm{x}_{io}$ includes individual one-hot encoding, observation one-hot encoding and observed features. The original design of FM as in Eq.~\eqref{eq1} embeds each component of the design feature vector into a latent vector. In contrast,  LMLFM (Eq.~\eqref{eq2})  embeds only the individuals and observations into latent vectors, and considers only the interactions among individuals, observations, and the design feature vector, thus simplifying the model. Instead of setting the latent dimension empirically as in the case of a FM, we initially let the latent dimension $k$ to be as large as the feature dimension $p$, and then apply variable selection to identify the relevant subset of latent factors (see below). This makes the proposed model almost as interpretable as a simple linear model while making use of factorization to accommodate nonlinear interactions among variables.

\subsection{Hierarchical Bayesian Model}
\label{sec:bayesian}
The hierarchical Bayesian structure of LMLFM is shown in 
Fig.~\ref{fig:LMLFM}. We decompose the task of joint selection of fixed and random effects into two sub-tasks: (i) identifying fixed and random effects by shrinking the variance of some variables towards 0; and (ii) selecting the relevant latent factors by shrinking some components towards 0. We handle the first sub-task by imposing a Laplace prior on  $\mu_k$ and $b_k$, respectively (see Laplace layer 2 in Fig.~\ref{fig:LMLFM}). For the second sub-task, we enforce sparsity in $\bm{\theta}_{\cdot k}$ by imposing Laplace prior on $\bm{\theta}_{\cdot k}$ (see Laplace layer 1 in Fig.~\ref{fig:LMLFM}). We denote the model parameters and hyper-priors of LMLFM by $\bm{\Theta}=\{\alpha,\Theta,\bm{\mu},\bm{b}\}$ and $\bm{\Theta}_0=\{\alpha_0,\beta_0,b_{\mu_0},b_{b_0}\}$ respectively, yielding the following generative model:
\resizebox{1.05\linewidth}{!}{
\begin{minipage}{\linewidth}
\begin{align*} \nonumber
&(y_{io}|\bm{x}_{io},\bm{\theta}_i,\bm{\theta}_o,\alpha) {\sim} N(y_{io}|\hat{y}_{io},\alpha^{-1}) & &(\alpha|\alpha_0,\beta_0) {\sim} Gamma(\alpha|\alpha_0,\beta_0) \\
&(\theta_{ik}|\mu_k^\mathcal{I},b_k^\mathcal{I}) {\sim} Laplace(\theta_{ik}|\mu_k^\mathcal{I},b_k^\mathcal{I}) & &(\theta_{ok}|\mu_k^\mathcal{O},b_k^\mathcal{O}) {\sim} Laplace(\theta_{ok}|\mu_k^\mathcal{O},b_k^\mathcal{O})\\
&(\mu_k^\mathcal{I}|b_{\mu_0}){\sim} Laplace(\mu_k^\mathcal{I}|0,b_{\mu_0}) & &(\mu_k^\mathcal{O}|b_{\mu_0}) {\sim} Laplace(\mu_k^\mathcal{O}|0,b_{\mu_0}) \\
&(b_k^\mathcal{I}|b_{b_0}) {\sim} Laplace(b_k^\mathcal{I}|0,b_{b_0}) && (b_k^\mathcal{O}|b_{b_0}) {\sim} Laplace(b_k^\mathcal{O}|0,b_{b_0})
\end{align*}
\end{minipage}
}
Unlike Bayesian FM  \cite{rendle2012factorization}, to accommodate different degrees of sparsity in relation to the numbers of individuals and observations, we allow different hierarchical priors for different latent factors for individuals  ($\Theta^{\mathcal{I}}$) and for observations ($\Theta^{\mathcal{O}}$) (See the grey nodes for  $\Theta^{\mathcal{I}}$ and black nodes for $\Theta^{\mathcal{O}}$ in Fig.~\ref{fig:LMLFM}).  We use $\bm{\mu}^\intercal=(\mu_k^\mathcal{I},\mu_k^\mathcal{O})_{k=1}^p,\bm{b}^\intercal=(b_k^\mathcal{I},b_k^\mathcal{O})_{k=1}^p$ to denote the mean and scale of the Laplace distribution respectively. The choice of the distribution of $y_{io}$ can be application-dependent. For the ease of exposition, we assume that the outcome variable follows a Gaussian distribution. 

We adopt the iterated conditional modes (ICM) algorithm \cite{besag1986statistical} to estimate the parameters of LMLFM. ICM updates blocks of parameters with the modes of their conditional posterior while keeping the remaining parameters fixed. Our choice of priors permits the analytical closed-form derivation of the modes of the conditional posterior density, yielding substantial speedup.
Specifically, we consider the Maximum A Posteriori (MAP) formulation:
\begin{equation}
\label{eq:map}
\argmax_{\bm{\Theta}}L=\pi\left(\bm{\Theta}|\bm{y},X,\bm{\Theta}_0\right)
\end{equation}
Due to space constraints, we include only the update equations for $\bm{\Theta}^\mathcal{I}=\left\{\alpha,\bm{\Theta}^\mathcal{I},\bm{\mu}^\mathcal{I},\bm{b}^\mathcal{I}\right\}$ here, omitting the superscript $\mathcal{I}$ to minimize notational clutter.

\para{Update of $\alpha$.}
The posterior of $\alpha$ is a Gamma distribution, whose mode is given by $\left(\beta_0 + \Norm{\bm{y}-\hat{\bm{y}}}_2^2/2\right)^{-1}\left( \alpha_0+\norm{\bm{y}}/2 -1 \right)$.

\para{Update of $\Theta$.}
For each model parameter $\bm{\theta}_i\in\Theta$, the prediction is a linear combination of two functions $g(i)$ and $h(i)$ that are independent of the value of $\bm{\theta}_i$:
\begin{equation}
\hat{\bm{y}}_i=g(i)+h(i)\bm{\theta}_i
\end{equation}
with $g(i) = \mathrm{diag}(X_i\cdot {\Theta_i^{\mathcal{O}}}^\intercal)$ and $h(i)=X_i + \Theta_i^\mathcal{O}$. Here $\Theta_i^\mathcal{O}$ is the matrix of latent factors constructed by the observations associated with $i$.  Hence, we have:
\begin{equation}
\theta_{ik}=\mu_k+\left(\bm{h}^\intercal_{ik}\bm{h}_{ik}\right)^{-1}\text{sgn}\left(\bm{r}_{ik}\right)\left(\norm{\bm{r}_{ik}}-1/\alpha b_k^\mathcal{I}\right)_+
\end{equation}
where $(\cdot)_+$ is the ReLU function; $\bm{r}_{ik} = \bm{h}^\intercal_{ik}(\bm{y}_i-g(i)-\sum_{q\in \{1:p\} \backslash k} \theta_{iq}\bm{h}_{iq}-\bm{h}_{ik}\mu_k)$, with $\{1:p\}\backslash k$ denoting the set of integers ranging from 1 to $p$ excluding $k$ and $\bm{h}_{ik}$ is the $k$-th column of $h(i)$. Clearly, sparsity is achieved if $\norm{\bm{r}_{ik}}\leq 1/\alpha b_k^\mathcal{I}$ and $\mu_k=0$.

\para{Update of $\bm{\mu}$.}
The problem of finding the optimal $\mu_k$ $(k=1,\cdots,p)$ reduces to finding the weighted median of the vector $\bm{\theta}_{\cdot k} \cup \{0\}$ with the weights $\{b_k\}_{i=1}^n \cup \{b_{\mu_0}\}$, yielding a linear-time algorithm \cite{gurwitz1990weighted}.

\para{Update of $\bm{b}$.}
The optimal $b_k$ $(k=1,\cdots,p)$ is updated by $2b_{b_0}\left( \sqrt{n^2+\frac{4}{b_{b_0}}\Norm{\bm{\theta}_{\cdot k}-\mu_k}_1 } -n \right)$.

We note that the computational complexity of LMLFM for one complete iteration is $O(\norm{S})$, which is linear in the size of the training data. Our approach is more efficient than FM \cite{rendle2012factorization}, whose computational complexity is $O(k\norm{S})$ ($k$ is the latent dimension). The space complexity of LMLFM is the same as that of FM, i.e., $O(\norm{\bm{y}})$.

\subsection{Effects and Outcome Estimation}
\label{param_est}
We proceed to describe how to estimate random effects, fixed effects and outcomes for seen and unseen individual/observation from the model.

\para{Temporal Individual-Specific Random Effects (TISE).}
As shown in Eq.~\eqref{eq2}, we can rewrite the prediction function of LMLFM in a form resembling ordinary linear regression where $\hat{y}_{io}=\bm{x}^\intercal_{io}{\bm{\gamma}}_{io}+\epsilon_{io}$ with coefficients ${\bm{\gamma}}_{io}=\bm{\theta}_i+\bm{\theta}_o$ and error $\epsilon_{io}=\bm{\theta}^\intercal_i \bm{\theta}_o$. Hence, we let ${\bm{\gamma}}_{io}$ be the estimator of TISE.

\para{Averaged Individual-Specific Random Effects (AISE).}
AISE is computed by integrating out the observation effects:
\begin{equation} \nonumber
\bm{\gamma}_i=\mathbb{E}_{\pi\left( \bm{\theta}_o | \bm{y},X,\bm{\Theta}_0 \right)}[\bm{\gamma}_{io}]=\bm{\theta}_i + \mathbb{E}_{\pi\left( \bm{\theta}_o | \bm{y},X,\bm{\Theta}_0 \right)}\left[\bm{\theta}_o\right]
\end{equation}
Solving $\mathbb{E}_{\pi\left( \bm{\theta}_o | \bm{y},X,\bm{\Theta}_0 \right)}\left[\bm{\theta}_o\right]$ is non-trivial. Hence, we  approximate it using the estimated observation factors, where $\mathbb{E}_{\pi\left( \bm{\theta}_o | \bm{y},X,\bm{\Theta}_0 \right)}\left[\bm{\theta}_o\right]\approx \frac{1}{m}\sum_{o\in\mathcal{O}}\bm{\theta}_o$.

\para{Temporal Population Averaged Random Effects (TPAE).}
Similar to solving AISE, TPAE is solved by integrating out the individual factors: ${\bm{\gamma}}_o=\mathbb{E}_{\pi\left( \bm{\theta}_i | \bm{y},X,\bm{\Theta}_0 \right)}[\bm{\gamma}_{io}]\approx \bm{\theta}_o + \frac{1}{n}\sum_{i\in\mathcal{I}} \bm{\theta}_i$.

\para{Fixed Effects.}
We say that a variable $k$ has fixed effect if and only if $b_k^{\mathcal{I}}=b_k^{\mathcal{O}}=0$. The fixed effect for variable $k$ is computed by $\beta_k=\mu_k^\mathcal{I}+\mu_k^\mathcal{O}$.

\para{Outcomes.}
The outcome $y_{io}$ for seen individual $i$ and observation $o$ is computed using Eq.~\eqref{eq2}. In the case of unseen individuals, we replace the posterior of the latent factors with their prior. Thus, for an unseen individual $i$, the outcome $y_{io}$ is given by:
\begin{equation} \nonumber
\mathbb{E}[y_{io}|\bm{x}_{io},\bm{\theta}_i,\bm{\theta}_o,\alpha] =\bm{x}_{io}^\intercal(\bm{\mu}^\mathcal{I}+\bm{\theta}_o)+\bm{\theta}^\intercal_o\bm{\mu}^\mathcal{I}
\end{equation}

\subsection{Convergence Analysis}
\label{sec:theoretical}
We establish two important properties in LMLFM: \textit{Ascent property} and \textit{Convergence}.\footnote{See supplemental material for detailed proofs.} Let $\mathbb{Z}^+$ denotes the set of all positive integers and $\theta^{(t)}$ denotes the value of $\theta$ at the $t$-th iteration. We denote $\pi\left(\theta|\right)$ as the full conditional posterior of $\theta$. 
\begin{prop}
\label{lemma3} \textbf{Ascent property.} $\pi\left(\bm{\Theta}^{(t+1)}|\right)\geq \pi\left(\bm{\Theta}^{(t)}|\right)$ holds for all iterations $t\in \mathbb{Z}^+$.
\end{prop}
The proof of Proposition~{\ref{lemma3}} follows from the observation that the joint posterior density of LMLFM is non-decreasing with the update of each component of $\bm{\Theta}$.
\begin{prop}
\label{lemma4} \textbf{Convergence.} If $\pi\left(\Theta^{(t)}|\right)$ is bounded above, there exists an iteration $t\in \mathbb{Z}^+$, such that $\forall i\in \mathbb{Z}^+$,$\norm{\pi\left(\Theta^{(t+i)}|\right) - \pi\left(\Theta^{(t)}|\right)} <\epsilon$ holds for $\epsilon>0$.
\end{prop}
We prove Proposition~\ref{lemma4} by contradiction, \ie the negation of the proposition implies that $\lim\limits_{t\to\infty}\pi\left(\Theta^{(t)}|\right) \to \infty$, yielding a contradiction.
	
\section{Experimental Evaluation}

\subsection{Experiments with Simulated Data}
\label{sec:simulation}
We report results of experiments with simulated data to answer the following questions: (RC1) Can LMLFM handle high-dimensional data? (RC2) Can LMLFM accurately select the relevant variables? (RC3) How does LMLFM perform in the presence of LC and CC?

\para{Simulated data.}
Following \cite{hui2017joint}, we construct simulated longitudinal data sets with $40$ individuals and $40$ observations per individual. We consider several choices of $p$ from $\{50,100,500,1000,5000\}$. We consider three types of correlation, \ie pure LC, pure CC and both (See supplemental material for details.).

\para{Methods compared.}
We compare LMLFM with several baseline methods: (i) State-of-the-art multi-level linear mixed model (M-LMM)  \cite{bates2014fitting}; (ii) State-of-the-art 2-level models: LMMLASSO, a linear mixed model with adaptive LASSO penalty on the fixed effects \cite{schelldorfer2011estimation}; GLMMLASSO, a generalized linear mixed model with standard LASSO penalty on the fixed effects \cite{groll2014variable} and rPQL \cite{hui2017joint}, a joint selection mixed model with adaptive LASSO penalty on the fixed effects and group LASSO penalty on the random effects; and (iii) Factorization-based multi-level Lasso (MLLASSO),\footnote{Despite its name, MLLASSO works only as a 2-level model and does not provide a simple way to associate the latent factors with random effects.} which factorizes the fixed effects as a product of global effects and individual effects,  both regularized by $\ell_1$ norm  \cite{lozano2012multi}; and (iv) the standard LASSO regression (LASSO) \cite{tibshirani1996regression}. We report performance statistics obtained from 100 independent runs. Hyper-parameters are selected using cross validation on the training data.
Evaluation scores are computed on the held-out data set. We report execution failure if an algorithm fails to converge within 48 hours or generates an execution error.

\begin{table}[tb]
\caption{Performance comparison on simulated data in the presence of multi-level correlation. We use `-' to denote execution failure.}
\label{tab:simulation}
\begin{center}
\resizebox{.95\columnwidth}{!}{%
\begin{tabular}{lllllll}
\toprule
\multirow{2}{*}{Method} & $p=100$ &      &      & $p=5000$ &      &      \\
\cmidrule{2-7}
                       & R$^2$ (\%)   & f.p. & f.n. & R$^2$ (\%)    & f.p. & f.n. \\
\midrule
LMLFM                   &   \bf{92}$\pm$1       &     \bf{0.2} &  0.8    &    \bf{88}$\pm$2      & 2.2     & 4.2     \\
rPQL                &   88$\pm$2     &  20.6    &  \bf{0}    &  -        & -     &  -    \\
M-LMM                   & 90$\pm$1     &   92   & \bf{0}     & -         & -     & -     \\
GLMMLASSO               &   83$\pm$4    & 91     & \bf{0}     & -         &  -    & -     \\
LMMLASSO                &   88$\pm$2        & 92     & \bf{0}     &  -        & -     &  -    \\
LASSO                 &   88$\pm$1     &  42.4    &  \bf{0}    & 84$\pm$4         &    415.8  & 0.4     \\
MLLASSO               &   40$\pm$8        & 23.8     & 0.8     &  1$\pm$1        &  \bf{0}    &  6.2    \\
\bottomrule
\end{tabular}
}
\end{center}
\end{table}

\para{Evaluation Measures.}
We evaluate the performance of all methods in terms of both predictive accuracy and the ability to select random and fixed effects. We measure the predictive accuracy using the r-squared ($\mathrm{R}^2$) score. To assess a method's ability to select the relevant variables, we consider a variable to be selected if the corresponding coefficient is non-zero. We use false positive (f.p.), the number of variables that are incorrectly selected, and false negative (f.n.), the number of variables that are incorrectly discarded, to assess the models.

\begin{figure*}[ht]
\centering 
\subfigure[SWAN data]{
\begin{minipage}[c]{0.47\textwidth} 
\centerline{\includegraphics[scale=0.4]{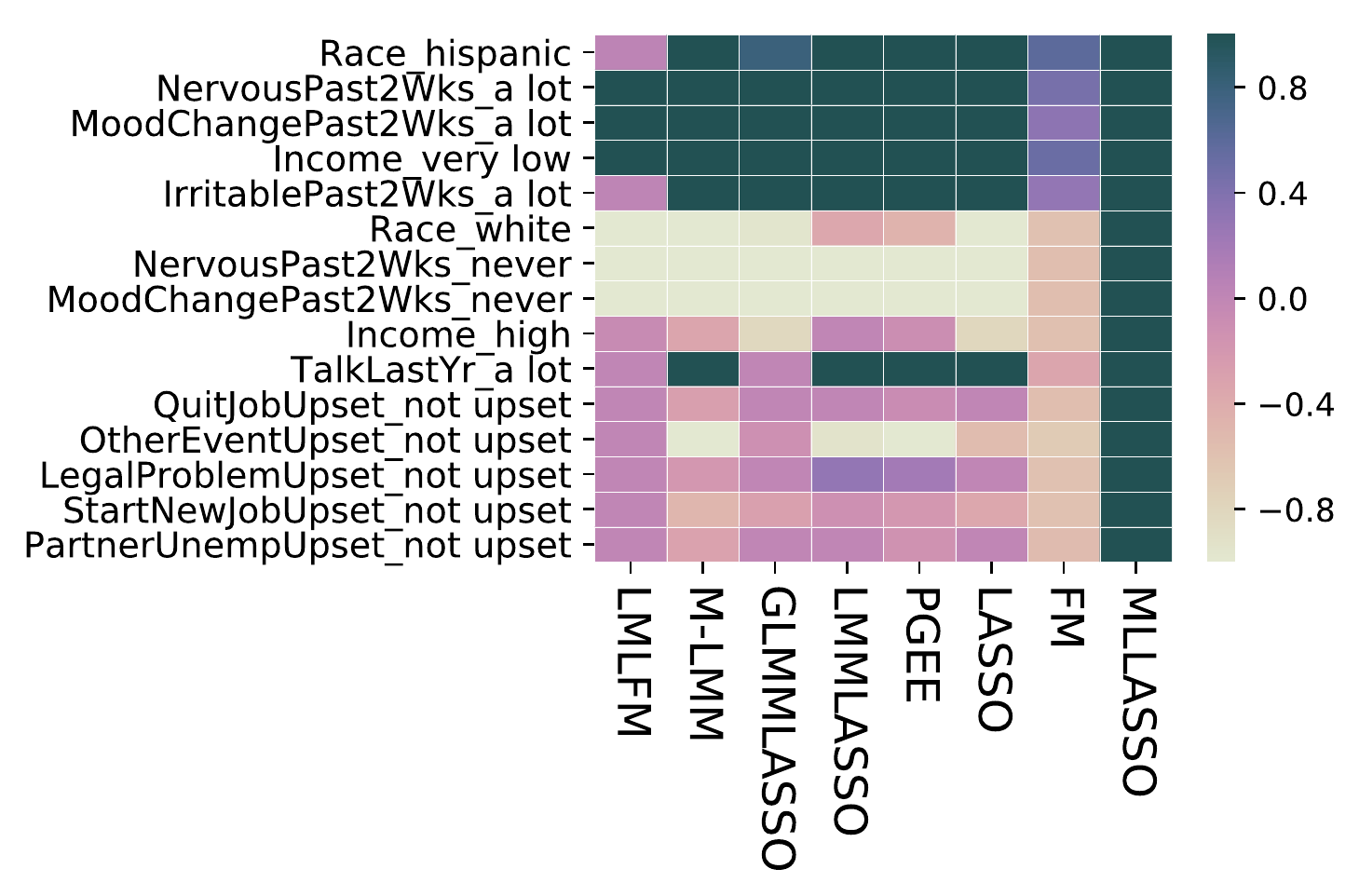}} 
\label{fig:swan_select} 
\end{minipage} 
}
\subfigure[GSS data]{
\begin{minipage}[c]{0.47\textwidth} 
\centering
\centerline{\includegraphics[scale=0.4]{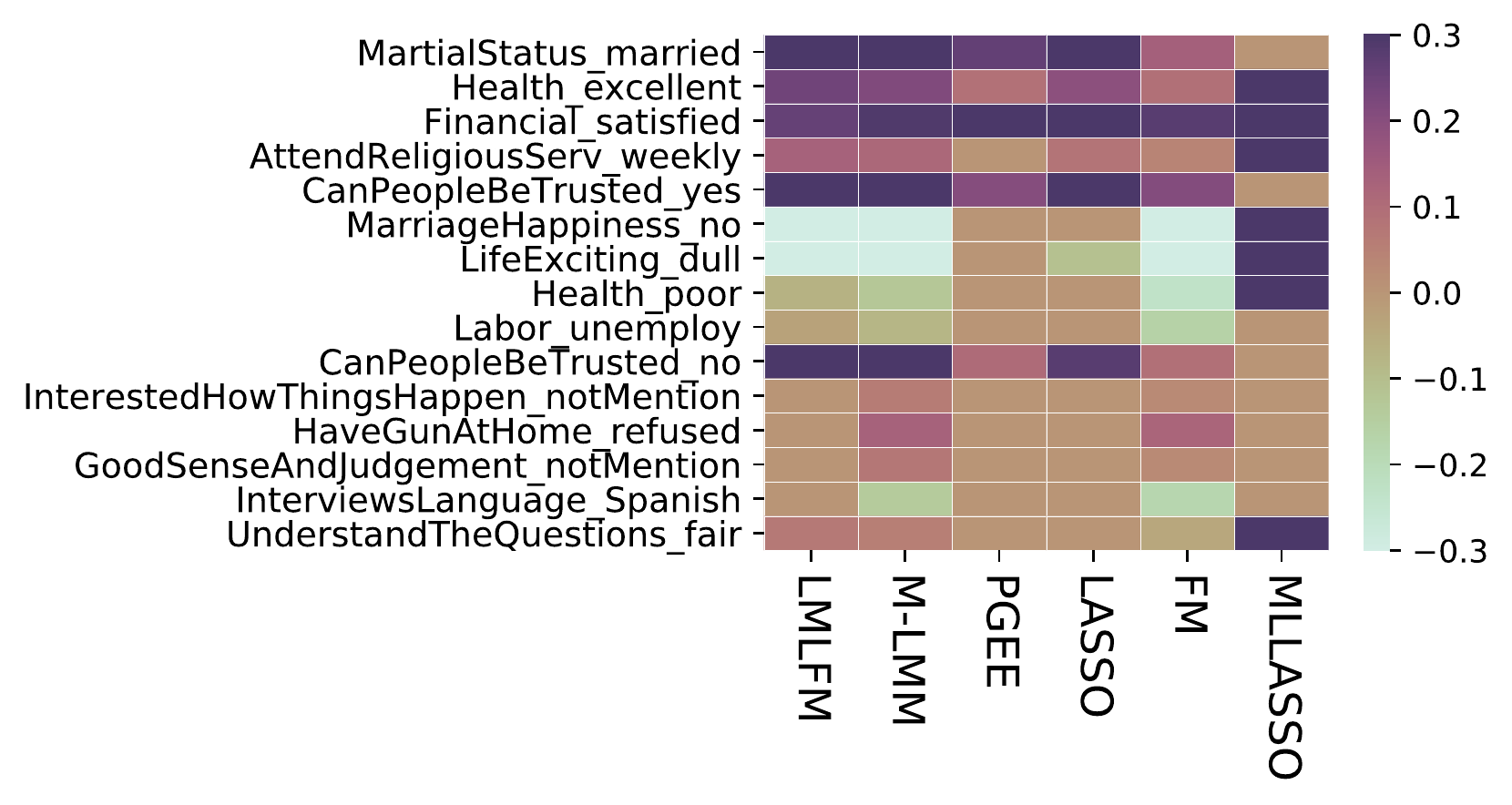}} 
\label{fig:gss_select} 
\end{minipage} 
}
\caption{Comparison of population averaged effects on the selected variables for SWAN and GSS data. The Top 5, middle 5 and bottom 5 variables have positive, negative and neutral effects respectively.}
\label{fig:fig3}
\end{figure*}

\para{Results.}
A subset of our  results summarized in Table~\ref{tab:simulation} answer RC1-RC3. RC1: the performance of the state-of-the-art mixed effects models are highly sensitive to the number of random effects, \ie M-LMM, LMMLASSO and GLMMLASSO fail when $p$ exceeds 100 due to execution error; Only LMLFM, LASSO and MLLASSO ran to completion. RC2: Selecting random effects is more challenging compared to selecting fixed effects. Among all models, only LMLFM and rPQL are designed to select random effects. To enforce model sparsity, say on variable $k$, LMLFM and rPQL shrink the vector $\bm{\theta}_{\cdot k}$ to 0 whereas other methods shrink a scalar $\beta_k$ to 0. Our results show that LMLFM  achieves the best terms of f.p. score. Although LASSO has attractive R$^2$ when $p=5000$,   LASSO is unable to select variables with random effects, leading to very high f.p. in the presence of LC, CC and both. RC3: Our results (omitted due to space constraints) show that 2-level models work poorly when a CC model is used on data with LC or vice versa. In contrast, multi-level models (M-LMM and LMLFM) achieve better fit on the data that exhibit LC, CC, or both. LMLFM consistently outperforms M-LMM in terms of accuracy, variable selection ability and computational efficiency. We conclude that LMLFM is the only method among those compared in this study, that can effectively model high-dimensional longitudinal data, and select the relevant variables, regardless of whether they are associated with random effects or fixed effects, in the presence of LC, CC, or both.

\subsection{Experiments with Real-World Data}
\label{sec:application}

\begin{table*}[tb]
\caption{Comparison of predictive accuracy on two real-life data sets. We use subscript `$-$' and `$+$' to denote the score on data with the 15 selected variables and all variables, respectively. We use `-' to denote execution failure.}
\label{tab:r2}
\begin{center}
\resizebox{.65\textwidth}{!}{%
\begin{tabular}{lllllllll}
\toprule
\multirow{2}{*}{Method}          & \multicolumn{4}{l}{SWAN} & \multicolumn{4}{l}{GSS} \\
\cmidrule{2-9}
& R$^2_- (\%)$   & f.p.$_-$   & f.n.$_-$ & R$^2_+ (\%)$  & R$^2_- (\%)$   & f.p.$_-$  & f.n.$_-$  & R$^2_+ (\%)$    \\
\midrule
LMLFM & \bf{30}$\pm$4       &  \bf{0}  &     \bf{0}            & \bf{49}$\pm$2                  & \bf{17}$\pm$2       & 2  & \bf{0}                & \bf{55}$\pm$2                        \\
M-LMM   & 29$\pm$2     & 1  & 5                   & -                  & 16$\pm$1     & 2 & 1                   & -                        \\
GLMMLASSO  & 19$\pm$3  & \bf{0}  & 2                     & -    & - & -                    & -                                                   & -                     \\
LMMLASSO &  26$\pm$3  & 2  & 2                     & -  & - & - 
 & -                                                          & -                      \\

PGEE    & 25$\pm$4   & 2 & 5                     & -                      & 3$\pm$1    & \bf{1} & \bf{0}                    & -                        \\

LASSO    & 21$\pm$4   & 1 & 2                     & 47$\pm$1                      & 0$\pm$1    & \bf{1} & \bf{0}                    & 47$\pm$1                        \\
FM       & 29$\pm$3    &\bf{0}  &5                    &  45$\pm$2        & 12$\pm$4  & 4  &2                       &31$\pm$2  \\
RF    & 23$\pm$5    &10  & \bf{0}                   & 47$\pm$1                 & 4$\pm$1        &10 & \bf{0}                   &                       \bf{55}$\pm$2 \\
MLLASSO    & 0$\pm$2         &10  & \bf{0}              &   20$\pm$2                  & -42$\pm$10       & 4 &\bf{0}                 &                       -2$\pm$1 \\
\bottomrule
\end{tabular}
}
\end{center}
\end{table*}

We compare LMLFM with the state-of-the-art baselines on two real-world longitudinal data sets: (i) Study of Women's Health Across the Nation (SWAN)  \footnote{\url{https://www.swanstudy.org/}} \cite{sutton2005sex} (on predicting depression); and (ii) General social survey (GSS)\footnote{\url{http://gss.norc.org/}} \cite{smith2017general} (on predicting general happiness). We chose these two data sets because they have attracted much interest in the field of social sciences.  We use the same settings of hyper-parameters for LMLFM as in our experiments with simulated data. We exclude rPQL because it fails on all of the experiments due to memory issue. In addition to the aforementioned baselines, we include some popular 1-level models in our comparison: Random Forest (RF), FM and Penalized GEE (PGEE) \cite{inan2017pgee}.

We seek answers to the following question: (RC4) How does LMLFM compare with the state-of-the-art baselines with respect to its ability to correctly identify the fixed and random effects and predictive accuracy? To answer RC4, for each data set, we choose as "ground truth", 5 "positive", 5 "negative" variables identified in the existing literature (see below for specifics) and add 5 additional variables that are believed to be relatively uninformative. 

\para{Evaluation on SWAN Data.}
In the case of SWAN data, we consider the task of predicting the CESD score \cite{dugan2015association}, which is used for screening for depression. The variables of interest include aspects of physical and mental health, and demographic factors, such as race and income. The data set includes 3,300 individuals, with 1-22 observations per individual, and 137 variables. The outcome we aim to predict is defined by $y_{io}=\mathrm{CESD}_{io}-15$ ($i$ is individual and $o$ is the age of the individual) since CESD $\geq 16$ has been observed to be highly indicative of depression. Existing research \cite{dugan2015association,prairie2015symptoms} suggests that hispanic ethnicity, depressed or fluctuating mood and low household income are highly positively correlated with depression, whereas Caucasian/white ethnicity, stable mood and high income are negatively correlated with depression. The variables used to answer RC4 and the experimental results are summarized in Fig.~\ref{fig:swan_select} and Table~\ref{tab:r2} respectively. We note that LMLFM outperforms all other methods in R$^2$ score and correctly recovers the relevant variables. Performance of the factorization baselines (FM, MLLASSO) is unsatisfactory. This is because of the lack of intuitive way to relate estimated latent factors to the corresponding effects. Note that the variables related to depressed mood are generally selected by our baselines (not shown), which is consistent with the findings in \cite{prairie2015symptoms}. FM renders menopausal status as strong factors to depression, a finding supported by existing literature \cite{bromberger2011mood,prairie2015symptoms}. However, we argue that depressed and fluctuated mood is more likely to be direct causes to depressive symptoms because menopausal status usually causes abnormal hormone level, which could further affect the mood of the patients. Results of LMLFM show that $\bm{\mu}^\mathcal{I}=\bm{0}$ and the sparsity rate of $\Theta^\mathcal{I}$ (\ie the number of zero components in $\Theta^\mathcal{I}$ divided by $\norm{\Theta^\mathcal{I}}$) is 98.3\%, which further implies that the random effects related to the individuals are less predictive. Nonetheless, the analysis on $\bm{\mu}^\mathcal{O}$ reveals a different story: 59 out of 136 $\bm{\mu^\mathcal{O}} $ are non-zero and the sparsity rate of $\Theta^\mathcal{O}$ is 56.4\%. This suggests that the depressive symptoms for multiple individuals with similar  age are correlated (CC) as are the depressive symptoms for a single individual across time (LC), with CC dominating LC. 

\para{Evaluation on GSS Data.}
In our experiments with the GSS data, we consider the problem of predicting the self-reported happiness. We define $y_{io}=1$ by individual $i$ reports happy at year $o$ and $y_{io}=-1$ as the opposite. The GSS data consists of 4,510 individuals, 1-30 observations per individual and 1,553 variables. Existing research \cite{dolan2008we,oishi2011income} indicates that, being married, good physical and psychological health, satisfactory with financial situation, having strong religious beliefs and being trusted are positively correlated with happiness, whereas the absence of these characteristics and unemployment are negatively correlated with happiness. The variables used to answer RC4 and the results of our experiments are shown in Fig.~\ref{fig:gss_select} and Table~\ref{tab:r2} respectively. Though we see that PGEE and LASSO have the lowest f.p., their R$^2$ is relatively low. They tend to shrink the negative effects to zero, and perform poorly even on the training set, which strongly suggests that they underfit the data. LMLFM is competitive with the best performing methods in recovering the relevant variables, while significantly outperforming them in  predictive performance. We note that variable selection with the GSS data is far more challenging than with the SWAN data because of the substantially larger number of variables and collinearity of the variables. The low R$^2$ of FM and MLLASSO indicate that they significantly underfit the data. Though RF has a high R$^2$ score, variables selected by RF are harder to explain compared to that of the other baselines (not shown). In contrast, LASSO achieves lower R$^2$, but selects variables that are consistent with those selected by LMLFM. We further find that all of the variables selected by LMLFM are consistent with the findings reported in \cite{dolan2008we}. We find that $\Theta^\mathcal{I}=0$, thus ruling out LC. This is perhaps explained by the huge gap between consecutive observations (the survey is taken once per one to three years) within which many unobserved factors could potentially affect subjective happiness. We find that the sparsity rate of $\Theta^\mathcal{O}$ is 84.9\%.  Our analysis of the fixed effects shows that 126 out of 1199 effects are non-zero and among which, only 6 features have absolute effects greater than 0.1, thus vast majority of variables are uninformative.

\para{Summary of Experimental Results.}
We conclude that LMLFM outperforms all the baselines and is the only multi-level mixed effects model that can reliably select variables associated with fixed as well as random effects from high-dimensional longitudinal data.

\section{Conclusions}
\label{sec:conclusion}
We have introduced  LMLFM, for predictive modeling from longitudinal data when the number of variables is large compared to the population size, the fixed and random effects are a priori unspecified, the interactions among variables are nonlinear, and the data exhibit complex correlation (LC, CC, or both). LMLFM, a natural generalization of FM to longitudinal data setting, adopts a novel hierarchical Bayesian model with two layers of Laplace prior, where the first layer induces a sparse latent representation and the second layer identifies fixed effects and random effects. We train LMLFM using iterated conditional modes algorithm which offers both computational efficiency and  strong convergence guarantee. Compared to the state-of-the-art alternatives, LMLFM yields more compact, easy-to-interpret, rapidly trainable, and hence scalable, models with minimal need for hyper-parameter tuning. Our experiments with simulated data with thousands of variables, and two widely studied real-world longitudinal data sets have shown that LMLFM outperforms the 1-level baselines and state-of-the-art 2-level and multi-level longitudinal models in terms of predictive accuracy, variable selection ability, and scalability to data with large number of variables. 


\subsection*{Acknowledgments}This work was funded in part by the  NIH NCATS through the grant UL1 TR002014 and by the NSF through the grants 1518732, 1640834, and 1636795, the Edward Frymoyer Endowed Professorship  at Pennsylvania State and the Sudha Murty Distinguished Visiting Chair in Neurocomputing and Data Science funded by the Pratiksha Trust at the Indian Institute of Science (both held by Vasant Honavar). The content is solely the responsibility of the authors and does not necessarily represent the official views of the sponsors.

\bibliographystyle{aaai}
\bibliography{AAAI-LiangJ.8186}
\end{document}